\pdfoutput=1
\documentclass[11pt]{article}

\usepackage[final]{acl}

\usepackage{multirow}
\usepackage{placeins}
\usepackage{times}
\usepackage{latexsym}
\usepackage{pbox}
\usepackage[T1]{fontenc}
\usepackage[utf8]{inputenc}

\usepackage{microtype}

\usepackage{inconsolata}

\usepackage{colortbl}
\usepackage{graphicx}
\usepackage{xspace}
\usepackage{tabularx}
\usepackage{tabulary}
\usepackage{amsmath}
\usepackage{booktabs}
\usepackage{listings}
\usepackage[most]{tcolorbox}
\usepackage{colortbl}
\usepackage{cuted}
\usepackage{xcolor}
\usepackage{kotex}
\usepackage{pifont}
\usepackage{dirtytalk}
\usepackage{algorithm}
\usepackage{algpseudocode}
\usepackage[utf8]{inputenc}
\usepackage[T1]{fontenc}
\usepackage[perpage]{footmisc}
\usepackage{hyperref}

\newtcolorbox{instructionsbox}[1][]{
  colframe=cyan!75!black,    % Frame color
  colback=green!5!white,     % Background color
  coltitle=black,            % Color of the title text
  title=#1,                  % Optional title
  rounded corners,           % Corner style
  boxrule=0.5mm,             % Frame thickness
  boxsep=5pt,                % Space between content and box
  toptitle=1mm,              % Space above the title
  bottomtitle=1mm,           % Space below the title
  left=0pt,                 % Left padding
  right=0pt,                % Right padding
  top=0pt,                   % Top padding
  bottom=0pt,                % Bottom padding
  fonttitle=\bfseries        % Font style for the title
}

\newcommand{\dataset}{\textsc{HRMCR}\xspace}

\title{Multi-Step Reasoning in Korean and the Emergent Mirage}
\author{Guijin Son{\textsuperscript{1,2,3}} \quad Hyunwoo Ko{\textsuperscript{1}} \quad 
\textbf{Dasol Choi}{\textsuperscript{2}} \\ \\ 
OneLineAI{\textsuperscript{1}} \quad Yonsei University{\textsuperscript{2}} \quad MODULABS{\textsuperscript{3}} \\
\texttt{spthsrbwls123@yonsei.ac.kr} \\ 
}

\begin{document}
\maketitle
\begin{abstract}
We introduce \textsc{HRMCR} (HAE-RAE Multi-Step Commonsense Reasoning), a benchmark designed to evaluate large language models’ ability to perform multi-step reasoning in culturally specific contexts, focusing on Korean. The questions are automatically generated via templates and algorithms, requiring LLMs to integrate Korean cultural knowledge into sequential reasoning steps. Consistent with prior observations on emergent abilities, our experiments reveal that models trained on fewer than \(2 \cdot 10^{25}\) training FLOPs struggle to solve any questions, showing near-zero performance. Beyond this threshold, performance improves sharply. 
State-of-the-art models (e.g., O1) still score under 50\%, underscoring the difficulty of our tasks. Notably, stepwise analysis suggests the observed emergent behavior may stem from compounding errors across multiple steps rather than reflecting a genuinely new capability. We publicly release the benchmark and commit to regularly updating the dataset to prevent contamination.
% This work introduces the \dataset (HAE-RAE Multi-Step Commonsense Reasoning) Benchmark, a collection of multi-step reasoning questions automatically generated using templates and algorithms. These questions require LLMs to recall diverse aspects of Korean culture and perform multiple reasoning steps to solve them. 
% Our experiments show that models pre-trained with less than \(2 \cdot 10^{25}\) training FLOPs generally fail to solve any questions, exhibiting near-zero performance. Even after a sharp increase in performance beyond this threshold, state-of-the-art (SOTA) LLMs continue to struggle, with O1, OpenAI's latest reasoning model, scoring below 50\%. Our multi-step design reveals an interesting observation: this so-called "emergent behavior" is observable only in the final stage of the solution, suggesting it may result from accumulated errors across multiple steps. To prevent contamination, we publicly release the question set but withhold the generation algorithm. Instead, we commit to regularly updating the question set to ensure the benchmark remains robust and reliable.
\end{abstract}

\section{Introduction}

Large language models (LLMs) have shown notable success in solving complex reasoning tasks across STEM~\citep{rein2023gpqa} and mathematics~\citep{cobbe2021training}, facilitated by methods such as chain-of-thought prompting~\citep{wei2022chain} and inference-time scaling~\citep{brown2024large}. However, it is unclear how beneficial these high-level problem-solving abilities—particularly for Olympiad-level math problems—are in real-world, everyday scenarios. Moreover, most existing reasoning benchmarks emphasize universal knowledge~\citep{hendrycks2021measuring, fang2024mathodyssey} that remains constant across different cultures and languages. While these benchmarks~\citep{ko2025understand} effectively measure a model’s general reasoning capabilities, they are less suited for evaluating the model’s proficiency in language-specific or culturally grounded reasoning.

% Large language models (LLMs), equipped with techniques like chain-of-thought prompting~\citep{wei2022chain} and inference-time scaling~\citep{brown2024large}, demonstrate impressive performance in solving complex reasoning problems in STEM~\citep{rein2023gpqa} and mathematics~\citep{cobbe2021training}. However, the practicality of solving Olympiad-level math problems—questions that the average user is unlikely to encounter—remains questionable regarding its usefulness for everyday applications. Furthermore, most reasoning benchmarks focus on culture-agnostic, universal knowledge that yields the same answers regardless of location. While effective for evaluating a model's general reasoning capabilities, such questions are less suitable for assessing an LLM's ability to reason in a specific language. 

\begin{figure}[t]
    \centering
    \includegraphics[width=1.0\linewidth]{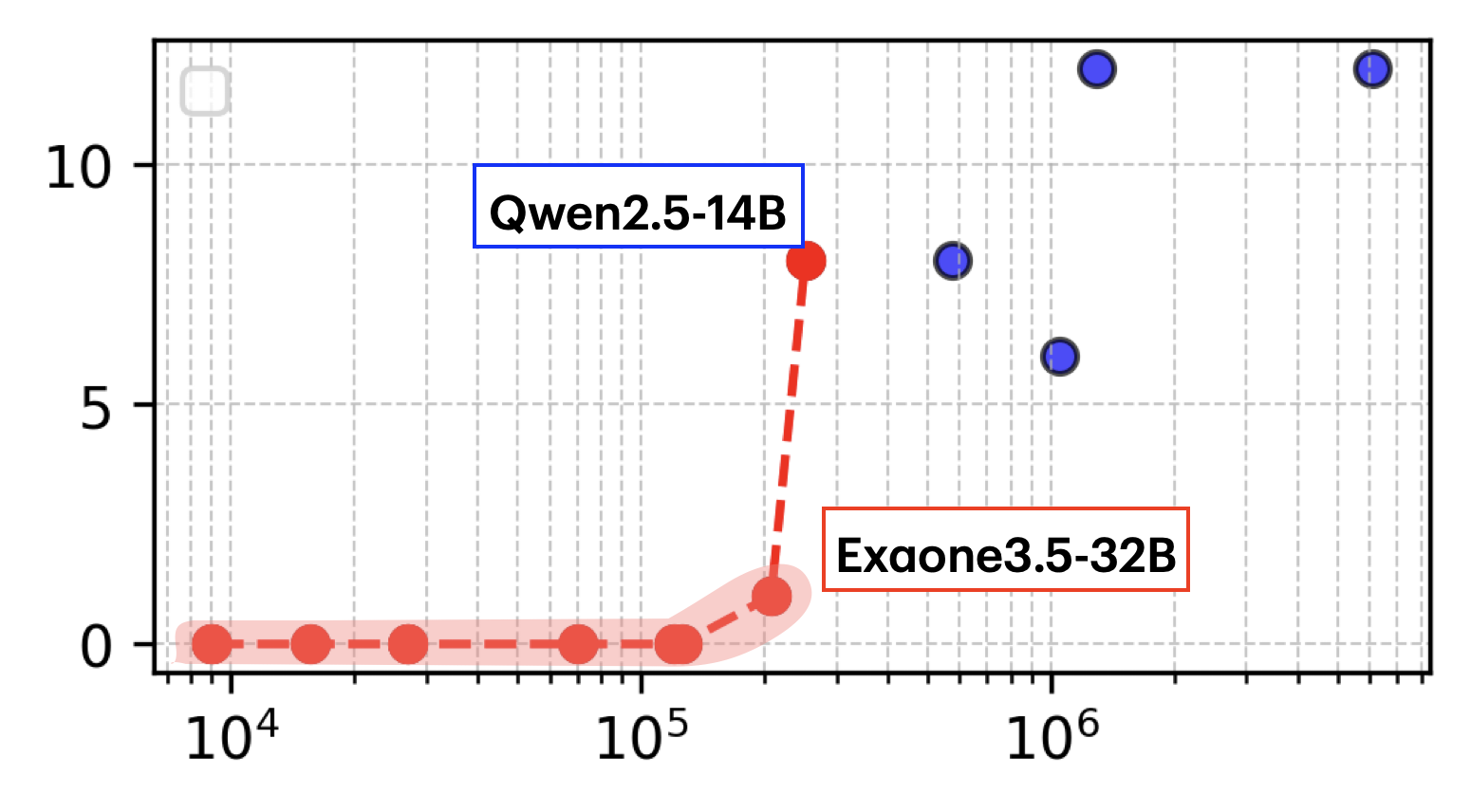}
    \caption{\footnotesize The X-axis represents the training compute scale in ExaFLOPs (10\(^ {18}\) floating-point operations), calculated as \(6 \times \text{\#parameters} \times \text{\#tokens}\) following \citet{kaplan2020scaling}. The Y-axis indicates the performance of the models on \dataset.}
    \label{fig:emergent}
    \vspace{-2mm}
\end{figure}

To address this gap, we introduce \dataset (HAE-RAE Multi-Step Commonsense Reasoning), a systematically generated benchmark that requires LLMs to integrate Korean cultural knowledge into multi-step reasoning. Our generation algorithm, which uses randomly selected seeds, generates questions and step-by-step solutions. Each of the two subsets in our benchmark comprises 50 questions. We make the \emph{question–answer set} publicly available but withhold the algorithm itself to prevent contamination and overfitting. Instead, we commit to regularly updating the benchmark\footnote{Link has been removed for anonymous submission. It will be publicly available.}.
% To this end, we introduce \dataset (HAE-RAE Multi-Step Commonsense Reasoning) Benchmark, a systematically generated collection of questions that require LLMs to incorporate Korean cultural knowledge and perform multiple reasoning steps. Our algorithm generates questions and step-by-step solutions based on a random seed. The benchmark consists of two subsets, each containing 50 questions. We publicly release the question set; however, we do not disclose the generation algorithm. Instead, we commit to regularly updating the question, preventing models from overfitting the benchmark and ensuring a more transparent evaluation.

As illustrated in Figure~\ref{fig:emergent}, models with fewer than \(2 \cdot 10^{25}\) training FLOPs struggle to solve any questions on the benchmark. Once training surpasses this threshold, however, scores increase sharply—a phenomenon often described as an emergent ability. Despite this improvement, even state-of-the-art (SOTA) LLMs (e.g., O1, GPT-4o, Claude-3.5-Sonnet) fall below 50\% accuracy, underscoring the benchmark’s difficulty. Notably, a stepwise performance analysis indicates that this apparent “emergence” may stem from cumulative errors across multiple reasoning steps, rather than a genuinely new capability.
% As shown in Figure~\ref{fig:emergent}, models trained with fewer than \(2 \cdot 10^{25}\) training FLOPs exhibit near-zero performance, followed by a sharp increase beyond this threshold. Conventionally, this could be interpreted as an "emergent ability." However, even state-of-the-art (SOTA) LLMs (e.g., O1, GPT-4o, Claude-3.5-Sonnet) score below 50\%, highlighting the benchmark's challenging nature. Interestingly, stepwise performance analysis reveals no evidence of emergence, suggesting that the observed emergence may instead result from cumulative errors across steps.

\begin{figure*}[h]
    \centering
    \includegraphics[width=\linewidth]{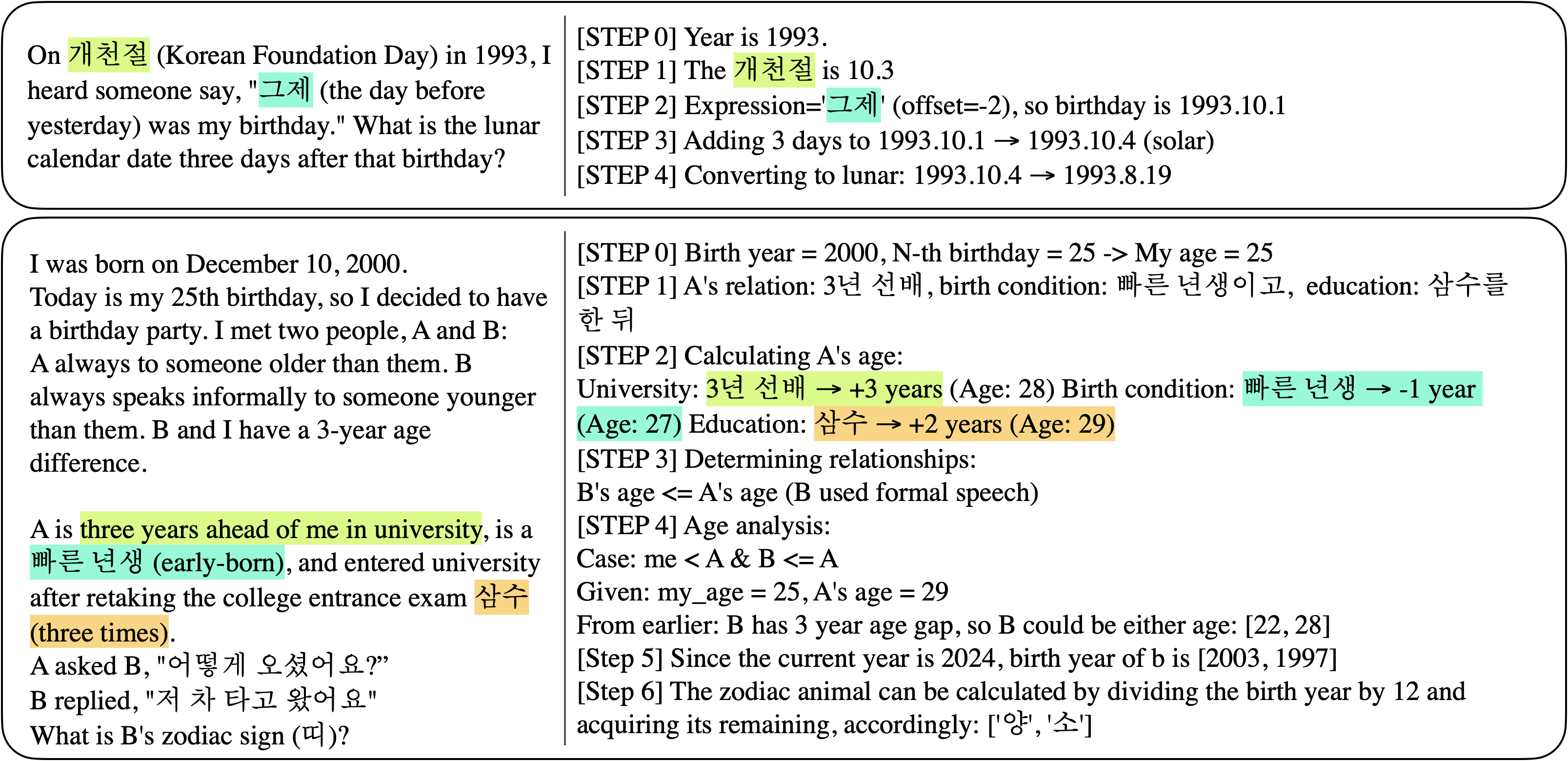}
    \caption{\footnotesize Example of generated questions in the \dataset benchmark. The figure showcases generated questions (left) alongside their automatically generated solutions (right). The top panel represents the "date" subset, while the bottom corresponds to the "zodiac" subset. Questions are translated into Korean to enhance accessibility.}
    \label{fig:hrmcr_overview}
\end{figure*}

\section{Related Works}
\paragraph{Korean Benchmarks}

Most publicly available Korean benchmarks primarily focus on knowledge~\citep{son2024kmmlu, kim2024click, son2023hae} or safety~\citep{lee2023square, lee2023kosbi}, leaving a gap in assessing more advanced reasoning skills. Consequently, the evaluation of Korean LLMs often relies on English benchmarks~\citep{research2024exaone3, research2024exaone3_5, yoo2024hyperclova}, to evaluate the reasoning capability of LLMs. This highlights the need for Korean-specific reasoning benchmarks that demand both linguistic and cultural competencies.
\paragraph{Multi-Step Reasoning}

As LLMs continue to enhance their reasoning abilities through improved pre-training~\citep{yang2024qwen2, lu2024mathcoder2} and post-training~\citep{wu2024inference}, many existing benchmarks no longer pose meaningful challenges or offer practical utility. Frontier LLMs~\citep{OpenAI2024o1} can now achieve near-expert performance on specialized tasks~\citep{rein2023gpqa, gao2024omni}, prompting the creation of even more demanding problem sets~\citep{glazer2024frontiermath}. However, those tasks often require domain-specific expertise—particularly in STEM—making them less representative of the broader range of reasoning tasks encountered in daily life.

One way to create challenging yet broadly accessible questions is to design multi-step problems that do not require deep expert knowledge. Such tasks can be generated by concatenating simpler subproblems~\citep{hosseini2024not, son2024multi} or by using template-based algorithms~\citep{sprague2023musr}, with the latter offering greater diversity. Building on prior work, we develop a fully algorithm-based benchmark that systematically generates multi-step reasoning problems demanding Korean commonsense knowledge. Additionally, we provide an automatic explanation generator, enabling more effective error analysis and evaluation for each step of the reasoning process.

\section{\dataset}

In this section, we introduce the \dataset (HAE-RAE Multi-Step Commonsense Reasoning) benchmark, describing its two subsets (Section~\ref{dataset_subset}) and key design choices (Section~\ref{dataset_design_choice}).

\subsection{Subsets}\label{dataset_subset}
\dataset comprises two subsets: \textbf{Date} and \textbf{Zodiac}, each created to require multiple steps of reasoning. We release 50 questions per subset.

\begin{algorithm*}
\caption{Question Generator for Date Subset}\label{algo_date}
\begin{algorithmic}[1]
\Require Database of cultural events with calendar types and dates
\Function{GenerateQuestion}{}
    \State \textit{// Step 1: Select base components}
    \State year, event $\gets$ Random(valid\_year\_range), RandomSelect(cultural\_events)
    \State cal\_type $\gets$ event.calendar\_type
    
    \State \textit{// Step 2: Select expressions}
    \State date\_expr, target\_cal $\gets$ RandomSelect(date\_expressions), RandomSelect(calendar\_types)
    
    \State \textit{// Step 3: Generate question}
    \State question $\gets$ Template( year, event.name, date\_expr, target\_cal)
    
    \Return question
\EndFunction
\end{algorithmic}
\end{algorithm*}

\paragraph{Date}
The Date subset contains concise, two-sentence questions involving Korean holidays and traditional date expressions. To solve these, the model must recall Korean cultural knowledge, perform basic arithmetic, and convert between solar and lunar calendars. Each question requires a five-step solution.

\paragraph{Zodiac}
The Zodiac subset features longer questions, typically spanning 10–12 lines. These tasks require handling a range of Korean cultural elements—such as the country’s unique age system\footnote{Korea calculates age differently from Western countries.}, conversational age expressions, and honorifics. The model must then extract logical inferences from the given premises and perform arithmetic to determine the zodiac sign. Each solution requires seven steps of reasoning at maximum.
% The Zodiac subset contains longer questions, each spanning 10 to 12 lines. To solve these, the model must recall concepts such as Korean age\footnote{Korea has a unique way of calculating age.} and various age expressions used in Korean conversations, understand honorifics, and make logical inferences based on the provided premises. Additionally, it must perform arithmetic to deduce the zodiac sign from the age. The gold response consists of six steps, making the Zodiac subset more challenging than the Date subset.

\subsection{Design Choice}\label{dataset_design_choice}
Each subset is generated by a dedicated algorithm comprising roughly 500 lines of code, and each algorithm includes a built-in solution generator that derives a gold-standard solution step-by-step. This built-in solver is a unique feature of our benchmark, as it facilitates comprehensive error analysis at each reasoning stage.

Benchmark contamination has become a pressing issue~\citep{xu2024benchmark, zhang2024careful}, and one common approach to address it is using private test sets~\citep{chollet2019measure}. However, private benchmarks have drawbacks: over the long term, repeated attempts can lead to overfitting, with models effectively using the test scores as a training signal~\citep{park2024open}. Additionally, keeping questions private hinders error analysis and limits opportunities for diagnosing and improving model performance. To balance these concerns, we publicly release our test set but keep the exact generation code confidential. If contamination arises, we can easily regenerate a new question–answer set by varying random seeds. By withholding the generation algorithm, we ensure that every newly released set remains unseen, making the benchmark more robust over time.

Algorithm~\ref{algo_date} illustrates the pseudo-code used to create the Date questions, and the algorithm for the Zodiac subset is provided in Appendix~\ref{app:model_details}. Figure~\ref{fig:hrmcr_overview} and \ref{fig:full_example} show examples of generated questions and their gold solutions.
% We provide a pseudo code of our generation process in Algorithm~\ref{algo_date} and \ref{algo_zodiac}. A sample of created questions and gold solutions are provided in Figure~\ref{fig:hrmcr_overview}.

\section{Experimental Setup}

In this section, we describe how responses were generated (Section~\ref{sec_resp_gen}) and evaluated (Section~\ref{sec_resp_eval}).
% In this section, we provide details on how the response was generated for each question (Section~\ref{sec_resp_gen}) and how the evaluation was done for each generated response (Section~\ref{sec_resp_eval}).

\subsection{Response Generation}\label{sec_resp_gen}
We evaluate a total of 20 LLMs, including proprietary models such as GPT-4o, GPT-4o-Mini~\citep{hurst2024gpt}, O1, O1-Mini~\citep{jaech2024openai}, and Claude-3.5-Sonnet~\citep{anthropic2024claude}, as well as open models like Qwen2.5~\citep{yang2024qwen2}, Llama3~\citep{dubey2024llama}, Exaone3.5~\citep{research2024exaone3_5}, and DeepSeek3~\citep{deepseekai2024deepseekv3technicalreport}. All models are evaluated in a greedy setting. For models up to 32B parameters, we run inference on a local GPU server; larger models are accessed via the OpenRouter API.\footnote{\url{https://openrouter.ai/}} For additional details on the evaluated models, see Appendix~\ref{app:model_details}.

% In this work, we evaluate a total of 20 LLMs, including proprietary models such as GPT-4o, GPT-4o-Mini~\citep{hurst2024gpt}, O1, O1-Mini~\citep{jaech2024openai}, and Claude-3.5-Sonnet~\citep{anthropic2024claude}, as well as open-source models like Qwen2.5~\citep{yang2024qwen2}, Llama3~\citep{dubey2024llama}, Exaone3.5~\citep{research2024exaone3_5}, and DeepSeek3~\citep{deepseekai2024deepseekv3technicalreport}. All models are evaluated in a greedy decoding setting. For models up to 32B, inference is performed on a local GPU server, while larger models are accessed via the OpenRouter API\footnote{\url{https://openrouter.ai/}}. For more details on the evaluated models see Appendix~\ref{app:model_details}.

\subsection{Response Evaluation}\label{sec_resp_eval}

\begin{figure*}[h]
    \centering
    \includegraphics[width=\linewidth]{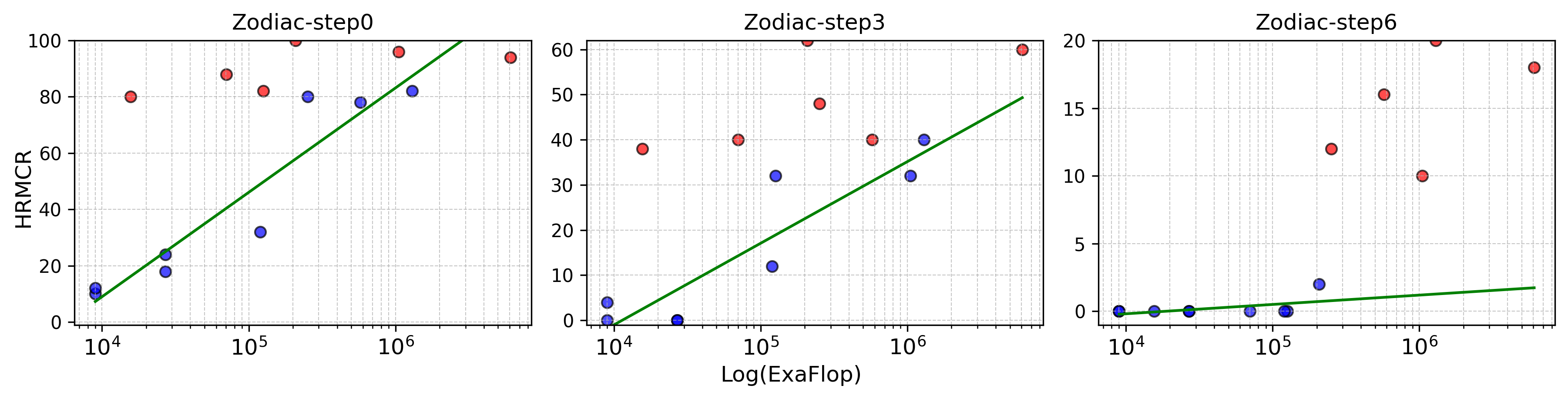}
    \caption{\footnotesize Breakdown of performance results for selected steps in the Zodiac subset. The green line represents the regression line, the blue points are instances used for fitting the regression, and the red points represent the test set. Only at the last step, the regression fails. For the entire results, see Appendix~\ref{app_additional_results}.}
    \label{fig:emergent_steps}
\end{figure*}

For evaluation, we use GPT-4o as an LLM-as-a-Judge~\citep{zheng2023judging}. The judge takes the question, the model-generated response, and the gold step-by-step solution. As shown in Figure~\ref{fig:judge_prompt} (Appendix~\ref{app:model_details}), the judge first provides a brief comparison with the gold solution and then determines whether the model’s response is correct. If it is incorrect, the judge identifies the specific step at which the error occurred. All evaluations use greedy decoding.

\section{Evaluation Results}

\begin{table}[ht]
\centering
\renewcommand{\arraystretch}{0.9}
\begin{tabular}{@{}lccc@{}}
\toprule
\multicolumn{1}{c}{\textbf{Models}} & \textbf{Date} & \textbf{Zodiac} & \textbf{Av.} \\
\midrule
o1 & \textbf{34} & \textbf{56} & \textbf{45} \\
GPT-4o & 28 & 32 & 30 \\
DeepSeek-V3 & 32 & 14 & 23 \\
Claude-3.5-Sonnet & \textbf{34} & 8 & 21 \\
Qwen2.5-72B & 4 & 20 & 12 \\
Llama3.1-405B & 6 & 18 & 12 \\
EXAONE3.5-32B & 0 & 2 & 1 \\ 
\bottomrule
\end{tabular}
\caption{\footnotesize Evaluation results on \dataset. We only display the performance of top-performing models per model family. The best-performing model is highlighted in \textbf{bold}.}
\label{tab_perf}
\vspace{-2mm}
\end{table}

Table~\ref{tab_perf} presents the evaluation results, from which we derive three key observations.
First, the \dataset benchmark is highly challenging: leading models such as GPT-4o, DeepSeek-V3, and Claude-3.5-Sonnet all score under 30\%. This is particularly noteworthy given that the benchmark is built on fixed, deterministic rules rather than specialized domain knowledge. Second, OpenAI’s latest reasoning-oriented LLM, O1, achieves an average score of 45, substantially outperforming earlier models. This suggests that inference-time scaling can generalize effectively to previously unseen domains. Finally, EXAONE3.5-32B, despite its size, shows near-zero performance on the benchmark. This indicates that solving \dataset requires not just model scale but also advanced training strategies and sufficient computational resources.
% Second, OpenAI’s latest reasoning LLM, O1, achieves an average score of 45, significantly outperforming all previous models. This indicates that inference-time scaling methods generalize effectively to unseen domains. Finally, EXAONE3.5-32B, a 32-billion-parameter model, demonstrates near-zero performance. This suggests that the ability to solve \dataset emerges only with extremely large computational resources and advanced capabilities.

\section{The Emergent Mirage}

\paragraph{Emergent at First Sight}~Emergent capability refers to abilities absent in smaller models but present in larger ones, making them unpredictable based solely on the performance of smaller models~\citep{wei2022emergent}. In Figure~\ref{fig:emergent}, we plot the performance of 19 models, with log compute on the X-axis and average performance on \dataset on the Y-axis. Models exhibit near-zero performance until reaching \(2 \cdot 10^{25}\) training FLOPs, followed by a sudden upsurge between Exaone3.5-32B and Qwen2.5-14B. This indicates that performance is driven primarily by training compute rather than model size: Qwen2.5-14B outperforms EXAONE3.5-32B by training on three times more tokens.
% This indicates that performance is driven not solely by model size but by training compute, as Qwen2.5-14B, despite being smaller, utilizes three times the token budget of Exaone3.5-32B.

\paragraph{Alternative Interpretation}\label{section_msr}
\citet{wei2022emergent} suggests that the sudden "emergence" of abilities may be attributed to multi-step reasoning. To investigate this hypothesis in our benchmark, we analyze the performance at each intermediate step. In Figure~\ref{fig:emergent_steps}, we plot step-wise accuracy and test for emergent behavior using linear regression. We train a regression model on smaller models (excluding the top five performers) to predict performance at each step. Surprisingly, while this regression accurately predicts the performance of larger models at all intermediate steps, it fails only at the final step. This pattern holds for both Date and Zodiac subsets, despite their final steps involving different types of operations. Given that the final step in the Zodiac subset involves basic arithmetic (simple division), which is not typically considered an emergent capability, we propose an alternative interpretation: rather than true emergence, this pattern may result from error accumulation across steps. We hypothesize that further decomposing tasks into smaller steps would reveal purely linear performance trends, challenging the notion of emergence in this context.
% In Figure~\ref{fig:emergent_steps}, we plot the accuracy for intermediate steps and test for emergent behavior. By definition, an "emergent capability" refers to performance that cannot be predicted from smaller models. To test this, we train a linear regression model excluding the top-performing five models. Interestingly, for both subsets, the regression accurately predicts the performance of the top models at all intermediate steps. However, at the final step of each task, the linear regression fails, indicating a sudden emergence of performance after surpassing a threshold compute budget. While this could be attributed to differences in step difficulty or coincidence—given that both subsets have a final step that appears emergent—this seems unlikely. For example, the last step of the Zodiac subset involves simple division, which is not typically considered emergent. We hypothesize that further dividing tasks into smaller steps would reveal fully linear trends across all steps, eliminating the appearance of emergent behavior at the final step.

\section{Conclusion}
We introduced HRMCR, a Korean multi-step reasoning benchmark that combines cultural knowledge with systematic generation. Our analysis revealed that seemingly emergent capabilities in LLMs may be artifacts of accumulated errors, prompting a reconsideration of how we evaluate model capabilities.

\section*{Acknowledgments}
This research was supported by Brian Impact Foundation, a non-profit organization dedicated to the advancement of science and technology for all. 
% % Bibliography entries for the entire Anthology, followed by custom entries
\bibliography{custom}
% % Custom bibliography entries only

\appendix

\begin{algorithm*}
\caption{Question Generator for Zodiac Subset}
\begin{algorithmic}[1]
\Require Database of relationships, speech levels, and age modifiers
\Ensure Question about age relationships and zodiac sign
\Function{GenerateQuestion}{}
    \State \textit{// Step 1: Generate base timeline}
    \State birth\_year, current\_year $\gets$ Random(valid\_range), Random(valid\_range)
    \State my\_age $\gets$ CalculateKoreanAge(birth\_year, current\_year)
    \State \textit{// Step 2: Generate person A's profile}
    \State relationship $\gets$ RandomSelect(university\_relationships)
    \State modifiers $\gets$ RandomSelect(age\_modifiers)
    \State a\_age $\gets$ CalculateAge(my\_age, relationship, modifiers)
    \State \textit{// Step 3: Generate conversation}
    \State speech\_level\_a, speech\_level\_b $\gets$ RandomSelect(speech\_levels),  RandomSelect(speech\_levels)
    \State age\_relationship $\gets$ DetermineRelationship(speech\_level\_a, speech\_level\_b)
    \State \textit{// Step 4: Generate question text}
    \State question $\gets$ Template(my\_age, relationship, modifiers, conversation)
    
    \Return question
\EndFunction
\end{algorithmic}
\label{algo_zodiac}
\end{algorithm*}

% \begin{table*}[t]
% \centering
% \fontsize{11}{13}\selectfont
% \begin{tabular}{lccc}
% \toprule
% \multicolumn{1}{c}{\textbf{Model}} & \textbf{Date} & \textbf{Zodiac} & \textbf{Avg.} \\
% \midrule
% \multicolumn{4}{l}{\textit{Proprietary models}} \\
% \midrule
% GPT-4o-Mini & - & - & 10 \\
% GPT-4o & - & - & 28 \\
% o1-mini & - & - & 0 \\
% o1 & - & - & 34 \\
% Claude-3.5-Sonnet & - & - & 34 \\
% \midrule
% \multicolumn{4}{l}{\textit{Open-source multilingual models}} \\
% \midrule
% Llama-3.2-1B & - & - & 0 \\
% Llama-3.2-3B & - & - & 0 \\
% Llama-3.1-8B & - & - & 0 \\
% Llama-3.3-70B & - & - & 2 \\
% Llama-3.1-405B & - & - & 6 \\
% Qwen2.5-0.5B & - & - & 0 \\
% Qwen2.5-1.5B & - & - & 0 \\
% Qwen2.5-7B & - & - & 0 \\
% Qwen2.5-14B & - & - & 4 \\
% Qwen2.5-32B & - & - & 0 \\
% Qwen2.5-72B & - & - & 4 \\
% Deepseek-V3 & - & - & 32 \\
% \midrule
% \multicolumn{4}{l}{\textit{English-Korean bilingual models}} \\
% \midrule
% EXAONE-3.5-2.4B & - & - & 0 \\
% EXAONE-3.5-7.8B & - & - & 0 \\
% EXAONE-3.5-32B & - & - & 0 \\
% \bottomrule
% \end{tabular}
% \caption{\footnotesize placeholder}\label{tab:performance}
% \end{table*}

\newpage

\section{Evaluated Models}\label{app:model_details}

\noindent\paragraph{Llama-3}~\citep{dubey2024llama}. The Llama-3 series, spanning from Llama-3 to Llama-3.3, comprises large language models ranging from 1 to 405 billion parameters developed by Meta. While these multilingual models are pretrained on datasets comprising 9 to 15 trillion tokens across multiple languages, they do not officially support Korean. The suite also provides instruction-tuned models.

\sloppy\paragraph{Qwen2.5}~\citep{yang2024qwen2}. Qwen2.5 is a suite of multilingual language models ranging from 0.5 to 72 billion parameters developed by Alibaba. Each model within the series is pretrained on a dataset of 18 trillion multilingual tokens, including Korean. Furthermore, the Qwen2.5 series officially support Korean and includes instruction-tuned variants for instruction-following.

\paragraph{DeepSeek-V3}~\citep{deepseekai2024deepseekv3technicalreport}. DeepSeek-V3 is a Mixture-of-Experts (MoE) language model with 671 billion parameters, utilizing 37 billion active parameters per token. It is trained on a dataset of 14.8 trillion multilingual tokens, making it robust across diverse languages and contexts. Additionally, they introduce the instruction-tuned version of the model.

\paragraph{EXAONE-3.5}~\citep{research2024exaone3_5}. EXAONE 3.5 is a suite of multilingual language models with sizes of 2.4B, 7.8B, and 32B parameters developed by LG AI Research. These models are pretrained on datasets comprising up to 9 trillion tokens, evenly balanced between Korean and English, to ensure strong bilingual capabilities. Only the instruction-tuned versions of the models were made publicly available.

\paragraph{GPT-4o \& o1}~\citep{hurst2024gpt, jaech2024openai}. GPT-4o, an advanced version of GPT-4, accepts multimodal inputs--including text, audio, image, and video--and demonstrate significant improvements on text in non-English languages. The o1 is a model designed to perform high-quality reasoning on complex tasks that require extensive thought processes, leveraging inference-time scaling through more elaborate reasoning steps and reflection. Both models are developed by OpenAI, and the latest versions of each were utilized in the experiments.

\paragraph{Claude-3.5}~\citep{anthropic2024claude}. Claude-3.5 series is the next generation of the Claude 3 series with significant improvements in commonsense and STEM reasoning benchmarks developed by Anthropic. We utilize the \texttt{claude-3.5-sonnet-20241022} for the experiments.

\section{Details in Evaluation}\label{app_additional_results}
In this section, we provide samples of the prompts used for evaluation, responses generated by GPT-4o as LLM-as-a-Judge, and additional evaluation results.

\begin{figure*}[h!]
\fontsize{8}{9}\selectfont
    \centering
    \begin{tabular}{|p{0.9\linewidth}|}
    \hline 
    \\
        \textbf{System Prompt:}
        \\
        Please act as an impartial judge and evaluate the quality of the response provided by an AI assistant to the provided question. You will be given a question, a gold step-by-step answer, and a response from an AI assistant. Review the response by the AI assistant. Compare it with the gold step-by-step answer. 
        \\\\
        Begin your evaluation by providing a comparison with the gold answer. Be as objective as possible. After providing your explanation, return whether the model has reached the correct answer, and if not specify which step it has first failed at. For example: \\\\
        if correct -> correct: [[true]] step: [[-1]]\\
        if wrong -> correct: [[false]] step: [[3]] \\
        \\
        \textbf{User Prompt:}\\
        \#\#\# Question : \\ \{question\} \\\\
        \#\#\# Gold Answer : \\ \{gold\} \\ \\
        \#\#\# Model Response : \\ \{response\}\\
\\\\
\hline
    \end{tabular}
    \caption{Prompt used for judging responses.}
    \label{fig:judge_prompt}
\end{figure*}

\subsection{Evaluation Prompt}
We use GPT-4o as an LLM-as-a-Judge for evaluation; in Figure~\ref{fig:judge_prompt}, we provide the prompt used for evaluation. The evaluation is done in greedy decoding settings.

\subsection{Sample Responses}
In Figure~\ref{fig:full_example}, we provide sample responses by models, and judgements generated by GPT-4o.

\begin{figure*}[h!]
\fontsize{8}{9}\selectfont
    \centering
    \begin{tabular}{|p{0.9\linewidth}|}
    \hline 
    \\
        \textbf{Question:}
        \\
        On 새해 첫날 (New Year's Day) in 1999, I heard someone say, "이튿날" (the day after tomorrow) was my birthday. What is the lunar calendar date 57 days after that birthday?
        \\\\
        \textbf{Gold Solution:}\\
        STEP 0: Year is 1999.\\
        STEP 1: The 새해 첫날 is 1.1 \\
        STEP 2: Expression='이튿날' (offset=+1), so birthday is 1999.1.2 \\
        STEP 3: Adding 57 days to 1999.1.2 → 1999.2.28 (solar) \\
        STEP 4: Converting to lunar: 1999.2.28 → 1999.1.13\\
        \\
        \textbf{Response by O1:}\\
        1999년 음력 설날(새해 첫날)은 \textbf{1월 1일}로, 그 다음 날인 \textbf{1월 2일}이 생일입니다. 이 생일로부터 57일 후의 음력 날짜를 계산해보겠습니다.\\\\
        1. \textbf{1월 2일부터 1월 말까지:} \\ 
        - 1월은 30일로 구성되어 있으므로, 1월 2일부터 1월 30일까지는 총 \textbf{29일}입니다.\\\\
        2. \textbf{2월:}\\
        - 1999년 음력 2월은 \textbf{29일}입니다.\\
        - 57일에서 1월에 사용된 29일을 제외하면, \textbf{28일}이 남습니다.\\
        - 따라서 2월 1일부터 2월 28일까지가 됩니다.\\\\
        결론적으로, \textbf{1월 2일}로부터 57일 후의 음력 날짜는 \textbf{2월 28일}입니다.\\\\
        \textbf{답:} 음력 2월 28일
        \\\\ 
        \textbf{LLM-as-a-Judge:}\\
        The AI assistant's response does not align with the gold step-by-step answer. Here's a comparison:\\\\
        1. \textbf{Year Identification:} \\
           - Both the gold answer and the AI response correctly identify the year as 1999.\\\\
        2. \textbf{New Year's Day Identification:} \\
           - The gold answer identifies the solar New Year's Day as January 1st, 1999, and the birthday as January 2nd, 1999.\\
           - The AI response incorrectly assumes the lunar New Year's Day as January 1st, 1999, and the birthday as January 2nd, 1999, without clarifying the calendar system.\\\\
        3. \textbf{Calculation of 57 Days After Birthday:} \\
           - The gold answer correctly calculates 57 days after January 2nd, 1999, in the solar calendar, resulting in February 28th, 1999.\\
           - The AI response incorrectly calculates the days using the lunar calendar, leading to a different conclusion.\\\\
        4. \textbf{Conversion to Lunar Date:} \\
           - The gold answer converts February 28th, 1999, from the solar calendar to the lunar calendar, resulting in the lunar date of January 13th, 1999.\\
           - The AI response does not perform this conversion and instead provides a lunar date of February 28th, which is incorrect.\\\\
        The AI assistant fails to correctly interpret the calendar system and does not perform the necessary conversion from the solar to the lunar calendar. The first failure occurs at step 3, where the AI incorrectly calculates the 57 days using the lunar calendar instead of the solar calendar.\\\\
        \textbf{Correct: [[false]] Step: [[3]]}
        \\\\
    \hline
    \end{tabular}
    \caption{A full example including a sample question, gold solution, model response (by o1) and judgment (by GPT-4o).}
    \label{fig:full_example}
\end{figure*}

\subsection{Additional Results}
\begin{figure*}[h]
    \centering
    \includegraphics[width=\linewidth]{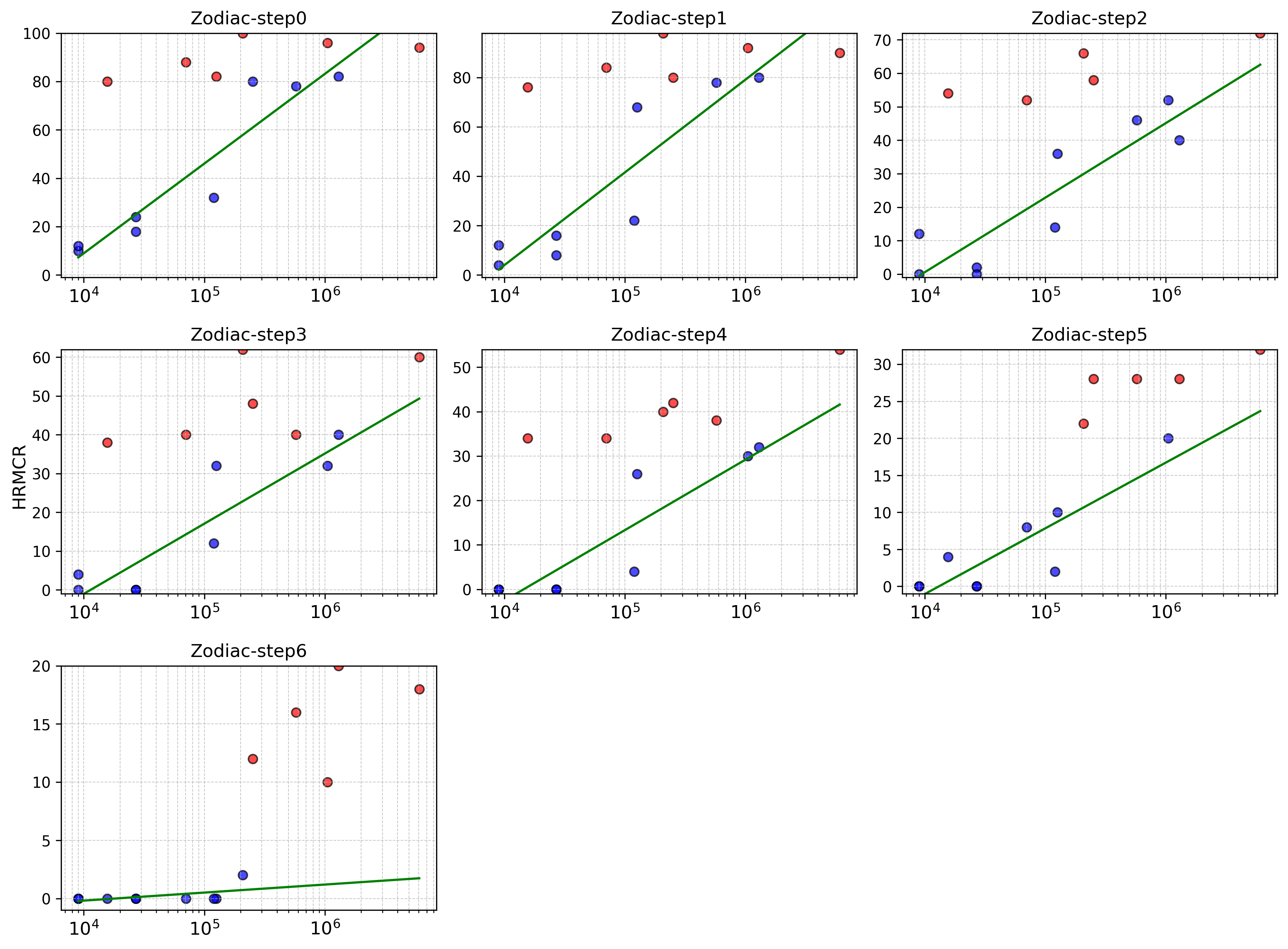}
    \caption{\footnotesize Breakdown of performance results for all steps in the Zodiac subset. The green line represents the regression line, the blue points are instances used for fitting the linear regression, and the red points represent the test set. Only at the final step the regression fails.}
    \label{fig:emergent_z_all}
\end{figure*}

\begin{figure*}[h]
    \centering
    \includegraphics[width=\linewidth]{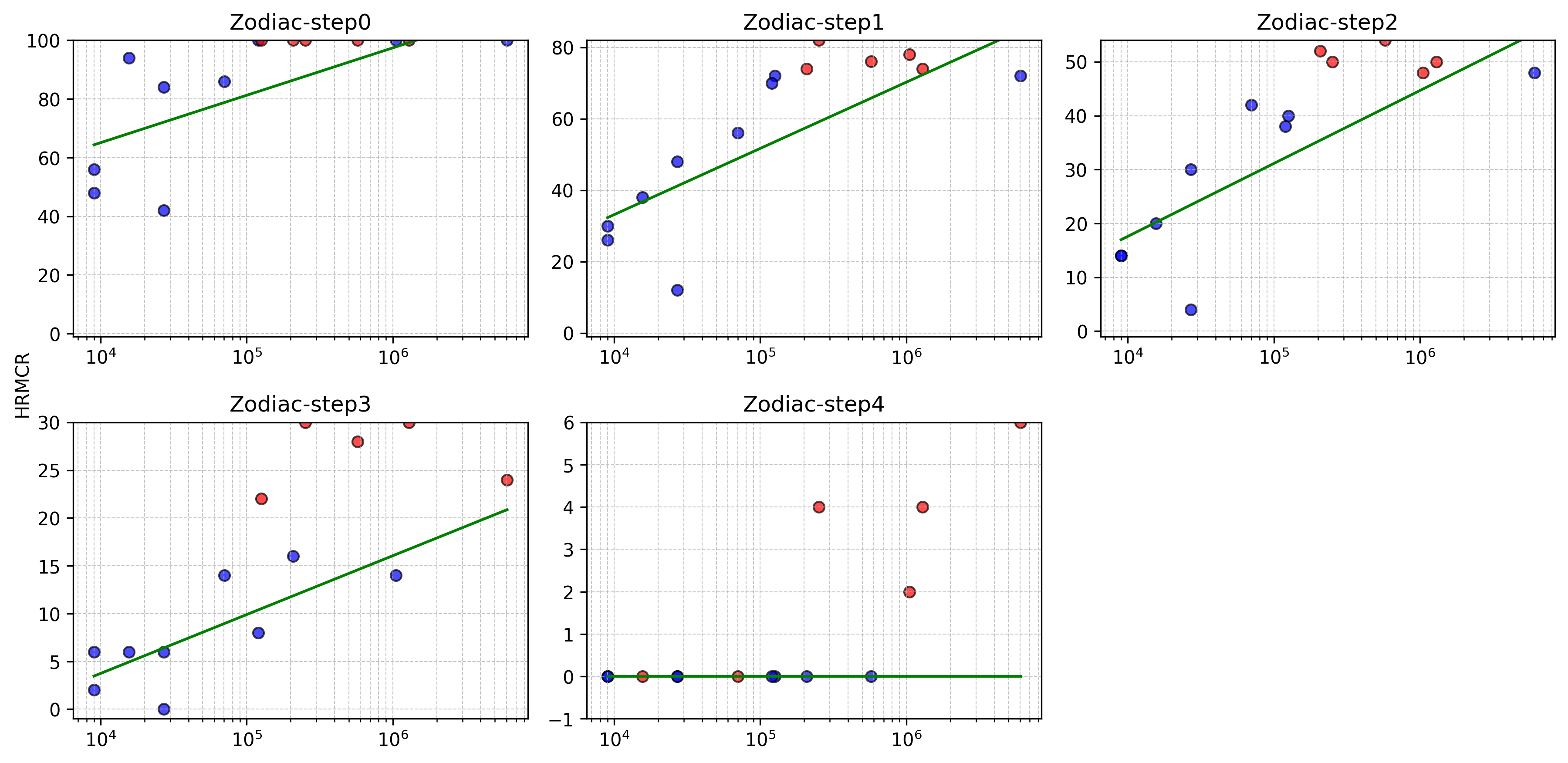}
    \caption{\footnotesize Breakdown of performance results for all steps in the Date subset. The green line represents the regression line, the blue points are instances used for fitting the linear regression, and the red points represent the test set. Only at the final step the regression fails.}
    \label{fig:emergent_d_all}
\end{figure*}

In this section, we present additional results. For the performance of all 20 models broken down by steps, see Table~\ref{tab_ent_1} and \ref{tab_ent_2}. In Figures~\ref{fig:emergent_d_all} and \ref{fig:emergent_z_all}, we provide continue from Section~\ref{section_msr} and provide regression results for all steps of both subsets. Our findings remain consistent with the additional results.

\begin{table*}[ht]
\centering
\renewcommand{\arraystretch}{0.9}
\begin{tabular}{lccccc}
\toprule
\textbf{Models} & \textbf{Step 0} & \textbf{Step 1} & \textbf{Step 2} & \textbf{Step 3} & \textbf{Final} \\
\midrule
Exaone3.5-2.4B & 94 & 38 & 20 & 6 & 0 \\
Exaone3.5-7.8B & 86 & 56 & 42 & 14 & 0 \\
Exaone3.5-32B & 100 & 74 & 52 & 16 & 0 \\
Qwen2.5-0.5B & 56 & 30 & 14 & 6 & 0 \\
Qwen2.5-1.5B & 42 & 12 & 4 & 0 & 0 \\
Qwen2.5-7B & 100 & 72 & 40 & 22 & 0 \\
Qwen2.5-14B & 100 & 82 & 50 & 30 & 4 \\
Qwen2.5-32B & 100 & 76 & 54 & 28 & 0 \\
Qwen2.5-72B & 100 & 74 & 50 & 30 & 4 \\
Llama3.2-1B & 48 & 26 & 14 & 2 & 0 \\
Llama3.2-3B & 84 & 48 & 30 & 6 & 0 \\
Llama3.1-8B & 100 & 70 & 38 & 8 & 0 \\
Llama-3.3-70B & 100 & 78 & 48 & 14 & 2 \\
Llama3.1-405B & 100 & 72 & 48 & 24 & 6 \\
DeepSeek3 & 100 & 70 & 54 & 40 & 32 \\
\midrule
Claude-3.5-Sonnet & 100 & 70 & 60 & 48 & 34 \\
GPT-4o-mini & 100 & 68 & 44 & 32 & 10 \\
GPT-4o & 100 & 76 & 54 & 40 & 28 \\
o1-mini & 100 & 76 & 46 & 32 & 0 \\
o1 & 100 & 76 & 64 & 60 & 34 \\ \bottomrule
\end{tabular}
\caption{\footnotesize Performance of all models on the Date subset, broken down per step.}
\label{tab_ent_1}
\end{table*}

\begin{table*}[ht]
\centering
\renewcommand{\arraystretch}{0.9}
\begin{tabular}{lccccccc}
\toprule
\textbf{Models} & \textbf{Step 0} & \textbf{Step 1} & \textbf{Step 2} & \textbf{Step 3} & \textbf{Step 4} & \textbf{Step 5} & \textbf{Final} \\
\midrule
Exaone3.5-2.4B & 80 & 76 & 54 & 38 & 34 & 4 & 0 \\
Exaone3.5-7.8B & 88 & 84 & 52 & 40 & 34 & 8 & 0 \\
Exaone3.5-32B & 100 & 98 & 66 & 62 & 40 & 22 & 2 \\
Qwen2.5-0.5B & 10 & 4 & 0 & 0 & 0 & 0 & 0 \\
Qwen2.5-1.5B & 24 & 16 & 2 & 0 & 0 & 0 & 0 \\
Qwen2.5-7B & 82 & 68 & 36 & 32 & 26 & 10 & 0 \\
Qwen2.5-14B & 80 & 80 & 58 & 48 & 42 & 28 & 12 \\
Qwen2.5-32B & 78 & 78 & 46 & 40 & 38 & 28 & 16 \\
Qwen2.5-72B & 82 & 80 & 40 & 40 & 32 & 28 & 20 \\
Llama3.2-1B & 12 & 12 & 12 & 4 & 0 & 0 & 0 \\
Llama3.2-3B & 18 & 8 & 0 & 0 & 0 & 0 & 0 \\
Llama3.1-8B & 32 & 22 & 14 & 12 & 4 & 2 & 0 \\
Llama-3.3-70B & 96 & 92 & 52 & 32 & 30 & 20 & 10 \\
Llama3.1-405B & 94 & 90 & 72 & 60 & 54 & 32 & 18 \\
DeepSeek3 & 94 & 94 & 70 & 60 & 48 & 20 & 14 \\
\midrule
Claude-3.5-Sonnet & 32 & 30 & 26 & 14 & 8 & 8 & 8 \\
GPT-4o-mini & 96 & 90 & 72 & 64 & 52 & 22 & 20 \\
GPT-4o & 88 & 84 & 54 & 50 & 42 & 32 & 32 \\
o1-mini & 92 & 90 & 70 & 64 & 48 & 44 & 44 \\
o1 & 98 & 98 & 72 & 68 & 58 & 56 & 56 \\ 
\bottomrule
\end{tabular}
\caption{\footnotesize Performance of all models on the Zodiac subset, broken down per step.}
\label{tab_ent_2}
\end{table*}

% \begin{figure*}[h]
%     \centering
%     \includegraphics[width=\linewidth]{figures/model_performance_date.pdf}
%     \caption{\footnotesize placeholder }
%     \label{fig:kmmlu_hr_result}
% \end{figure*}

% \section{Additional Resources for the \dataset dataset}
% \begin{figure*}[h]
%     \centering
%     \includegraphics[width=\linewidth]{figures/model_performance_zodiac.pdf}
%     \caption{\footnotesize placeholder }
%     \label{fig:kmmlu_hr_result}
% \end{figure*}

\end{document}